\documentclass{article} % For LaTeX2e
\usepackage{iclr2027_conference_preprint,times} % iclr2027_conference

\usepackage{amsmath,amsfonts,bm}

\def\eqref#1{equation~\ref{#1}}
\def\1{\bm{1}}

\DeclareMathAlphabet{\mathsfit}{\encodingdefault}{\sfdefault}{m}{sl}
\SetMathAlphabet{\mathsfit}{bold}{\encodingdefault}{\sfdefault}{bx}{n}

\usepackage{xcolor}   % load before hyperref so the colours below are available
\usepackage{hyperref}
\usepackage{url}

\usepackage{algorithm,algorithmic}
\usepackage{wrapfig,epsfig}
\usepackage{mathtools}
\usepackage{dsfont}
\usepackage{graphicx}
\usepackage{nomencl}
\usepackage{multirow}
\usepackage{makecell}
\usepackage{adjustbox}
\usepackage{subcaption}
\usepackage{enumerate}
\usepackage{diagbox}
\usepackage{enumitem}
\usepackage{multirow}
\usepackage{amssymb}
\usepackage[symbol]{footmisc}

\usepackage{booktabs}
\usepackage{bbding}
\usepackage{pifont}
\usepackage{wasysym}
\usepackage{utfsym}
\usepackage{fontawesome}
\usepackage{colortbl}
\usepackage{color}
\usepackage[dvipsnames]{xcolor}

\definecolor{citecolor}{HTML}{0B7285} % \citep / \citet  (teal)
\definecolor{linkcolor}{HTML}{1A0DAB} % internal refs: sections, figures, tables
\definecolor{urlcolor}{HTML}{1565C0}  % external URLs

\hypersetup{
    colorlinks=true,
    citecolor=citecolor,
    linkcolor=linkcolor,
    urlcolor=urlcolor
}

\newcommand{\best}[1]{\textbf{#1}}
\newcommand{\collapsed}[1]{\textcolor{gray}{#1}}

\newcommand{\na}{n/a}
\newcommand{\hjepa}{Human-JEPA}

\title{\hjepa: A Human-Centric Vision Model that Perceives and Anticipates}

\author{Hui Wei$^{1,2}$, Licai Sun$^{1,2}$, Guoying Zhao$^{1,2}$~\thanks{Corresponding author.} \\
$^1$ELLIS Institute Finland, Finland\\
$^2$Center for Machine Vision and Signal Analysis (CMVS), University of Oulu, Finland\\
\texttt{\{hui.wei, licai.sun, guoying.zhao\}@oulu.fi}\\
}

\iclrfinalcopy % Uncomment for camera-ready version, but NOT for submission.
\begin{document}

\maketitle

\begin{abstract}
Machines that understand humans should perceive the present and anticipate the future. 
Existing human-centric vision model are pretrained on human images, set the state of the art in static dense perception, so motion and anticipation are out of reach. 
Here we present \hjepa{}, a human-centric vision model trained on video by anchored forecasting: dense targets are pinned to a frozen copy of the initialization, preventing a silent collapse of dense perception, and block masks are replaced by a pure past-to-future split, avoiding a five-point action tax and a seventeen-point re-identification collapse.
Under frozen probes, Human-JEPA leads the pixel-anchored specialists on pose and person re-identification at 2.7 times fewer parameters, conceding high-resolution dense parsing, and its released predictor head is the first that does not degrade anticipation.
A single safely adapted model thus serves both halves of understanding humans.
\end{abstract}

\section{Introduction}
The human visual system perceives and anticipates simultaneously~\citep{rao1999predictive,summerfield2006predictive}. 
Watching someone reach across a table, we register the body, the pose, and who the person is, and at the same time we predict where the hand will land, well before it arrives. 
Machine vision for humans needs both halves. 
Perception of the present segments bodies~\citep{zhang2025deep}, localizes joints~\citep{zheng2023deep}, decides whether two views show the same person~\citep{shahreza2025foundation}, etc.
Anticipation of the future matters just as much, because decisions are made before an action completes~\citep{zatsarynna2025manta}. 

Existing strongest human-centric vision model comes from pixel-anchored image encoders pretrained on human corpora: Sapiens2~\citep{khirodkar2026sapiens} sets the state of the art in parsing and pose at high resolution, HAP~\citep{yuan2023hap} shows that body-part priors improve masked person modeling, and supervised generalists such as UniHCP~\citep{ci2023unihcp} and Hulk~\citep{wang2025hulk} unify human tasks behind shared heads. 
However, these models that consume human images can be the strongest possible reader of a photograph and still represent no motion and predict no future, so the second half of understanding humans is unreachable.

To address the limitation, we propose \hjepa{}, the first human-centric vision model that perceives and anticipates simultaneously. \hjepa{} built from the released V-JEPA~2.1~\citep{mur2026v} by \emph{anchored forecasting} with two changes and one omission.
The anchor pins dense context targets to a frozen copy of the initialization, removing the target drift that causes the collapse; an image co-training branch complements it and turns out to be where the identity ability originates.
The forecasting mask family replaces block inpainting with a pure past-to-future split, so the only route to the target is a model of how the scene evolves; this specializes the temporal representation to humans at almost no cost and trains the predictor as a same-space forecaster.

\begin{figure}[t]
\centering
\includegraphics[width=\linewidth]{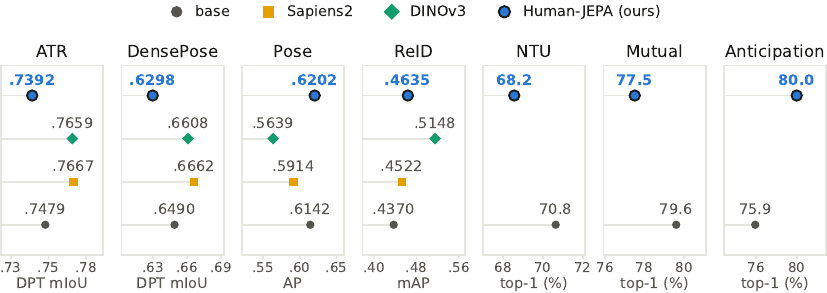}
\vspace{-6mm}
\caption{\textbf{What human specialization buys, and what it costs.} Seven frozen-probe tracks under one protocol. \hjepa{} (blue, outlined) leads the human specialists on pose, identity, and anticipates.}
\label{fig:teaser}
\vspace{-4mm}
\end{figure}

The experimental results, measured under frozen probes with an identical protocol per track (Figure~\ref{fig:teaser}), leads on every axis the image specialists cannot contest and on two they can.
Pose reaches $0.620$ AP over two pretraining seeds, ahead of every pixel-anchored specialist including the $2.7\times$ larger Sapiens2-0.8B by $2.9$ AP, of which the base family position contributes $2.3$ and the specialization $0.6$, an attribution we keep explicit throughout.
Person re-identification reaches $0.4635$ mAP on Market-1501, which turns the base's deficit against Sapiens2 into a lead and gains $2.7$ mAP over the base itself.
Anticipation gains $4.1$ points over the base's released bundle, of which $1.0$ comes from our encoder while the base's own shipped head costs it $3.0$. Overall, the key contributions of \hjepa{} are as follows:

\begin{itemize}[leftmargin=5mm]
\item We propose \textbf{\hjepa{}}, the first human-centric vision model that perceives and anticipates simultaneously: under frozen probes it leads the pixel-anchored specialists on pose and person re-identification at $2.7\times$ fewer parameters, and its predictor head does not degrade anticipation.
\item We develop an \textbf{anchored forecasting} that specializes a video JEPA without destroying it: a frozen-initialization dense anchor with an image co-training branch prevents the silent collapse of continued pretraining, and a past-to-future masking removes the action and identity tax.
\item We conduct \textbf{systematic evaluations} under one frozen-probe protocol, including a causal partner-ablation probe that falsifies all nine person-level objectives we construct.%, a quantified account of every cost, and a full release of weights, predictor, probes, and manifests.
\end{itemize}

\section{Related Work}

\subsection{Human-Centric Vision Models}
Human-centric vision has gradually moved from task-specific designs toward general-purpose representations that can be shared across diverse tasks~\citep{tang2025human}. Early unified models, such as UniHCP~\citep{ci2023unihcp} and PATH~\citep{tang2023humanbench}, jointly train a single Vision Transformer (ViT)~\citep{dosovitskiy2021an} across multiple human-centric datasets with task-specific queries or projectors, demonstrating that shared body semantics can benefit pose estimation, human parsing, re-identification, etc. 
Hulk~\citep{wang2025hulk} further extends this direction with a multimodal generalist framework that formulates heterogeneous 2D, 3D, skeleton-based, and vision-language tasks as translations between discrete and continuous output spaces, reducing the need for task-specific heads. 
In parallel, large-scale self-supervised pretraining has been explored to build high-fidelity human-centric backbones. Sapiens~\citep{khirodkar2024sapiens} pretrains masked autoencoders on hundreds of millions of in-the-wild human images and achieves strong dense prediction performance at native 1K resolution, while Sapiens2~\citep{khirodkar2026sapiens} further combines masked reconstruction with a self-distilled contrastive objective to preserve both low-level appearance cues and high-level semantics, scaling to billion-parameter models and 4K-resolution inputs. 
All of these operate on single images or labeled multi-task data; none represents how people move or interact over time, which is the axis \hjepa{} targets.

\subsection{Self-Supervised Learning}
Self-supervised learning (SSL) has emerged as a powerful paradigm for learning transferable representations from massive unlabeled data~\citep{gui2024survey}. Existing SSL methods can be broadly grouped into three paradigms: contrastive learning, distillation, and masked data modeling.

{The contrastive line} can be traced back to instance discrimination~\citep{wu2018unsupervised}, which treats each image as an individual class and learns representations by separating each instance from others using a memory bank. Building on this idea, CMC~\citep{tian2020contrastive} maximizes mutual information across multiple views of the same scene to encourage view-invariant representations. SimCLR~\citep{chen2020simple} further shows that strong augmentations, a nonlinear projection head, and large in-batch negatives are crucial for effective representation learning. To reduce the dependence on large batch sizes, MoCo~\citep{he2020momentum} introduces a queue-based dynamic dictionary updated by a momentum encoder. MoCov2~\citep{chen2020improved} further improves this framework with stronger augmentations and an MLP projection head, and MoCov3~\citep{chen2021empirical} extends it to ViT and addresses the training instability encountered in this setting.

Beyond explicit negative pairs, {distillation-based methods} learn representations by matching the outputs of two networks. BYOL~\citep{grill2020bootstrap} trains an online network to predict a momentum target network using only positive pairs, while SimSiam~\citep{chen2021exploring} shows that a simple Siamese framework with stop-gradient can avoid collapse without negatives, momentum encoders, or large batches. Along this direction, DINO~\citep{caron2021emerging} aligns the output distributions of a student and a momentum teacher through centering and sharpening, yielding ViT attention maps with emergent object-level segmentation. DINOv2~\citep{oquab2024dinov2} scales this recipe with curated data and optimized training to obtain general-purpose frozen features, and DINOv3~\citep{simeoni2025dinov3} further improves dense representations by scaling model and data and introducing a Gram-anchoring objective.

Finally, {masked data modeling} learns representations by reconstructing partially corrupted inputs. MAE~\citep{he2022masked} masks a large proportion of image patches and reconstructs the missing parts with an asymmetric encoder--decoder architecture, enabling scalable visual pretraining. Data2vec~\citep{baevski2022data2vec} provides a unified self-supervised framework for speech, vision, and language by predicting latent representations of the full input from a masked view, and the subsequent data2vec~2.0~\citep{baevski2023efficient} substantially improves the efficiency of this paradigm. 

\subsection{Joint Embedding Predictive Architecture}
Joint Embedding Predictive Architecture (JEPA) learns representations by predicting target-region embeddings from context embeddings in latent space, avoiding both augmentation-dependent positive pairs in contrastive learning and low-level biases of pixel/token reconstruction. I-JEPA~\citep{assran2023self} first demonstrated this paradigm for images, and subsequent variants extended it to video (V-JEPA~\citep{bardesvVJEPA}, V-JEPA~2~\citep{assran2025v}, V-JEPA~2.1~\citep{mur2026v}), motion (MC-JEPA~\citep{bardes2023mc}), point clouds and 3D scenes (Point-JEPA~\citep{saito2025point}, 3D-JEPA~\citep{hu20243d}), LiDAR (AD-L-JEPA~\citep{zhu2026self}), and audio (Audio-JEPA~\citep{tuncay2025audio}). Recent work further simplifies anti-collapse objectives with distributional regularization (LeJEPA~\citep{balestriero2025lejepa}, LeWorldModel~\citep{maes2026leworldmodel}) and introduces structured semantic priors through object-level masking or vision-language guidance (Causal-JEPA~\citep{nam2026causal}, ThinkJEPA~\citep{zhang2026thinkjepa}). These advances establish JEPAs as a general framework for learning structured representations across modalities and tasks. 
For human-centric learning, S-JEPA~\citep{abdelfattah2024s} applies latent prediction to 3D skeletons for action recognition.

\section{\hjepa{}}
\hjepa{} takes a released video JEPA, an encoder $f$ and a predictor $g$ pretrained on general internet video~\citep{mur2026v}, and returns a human-centric model of the same architecture and size, so that adopting it is a checkpoint swap for an existing user. The input is a curated corpus of human video and person crops, the output is a frozen-transferable encoder together with a predictor trained in that encoder's own representation space, and the objective is self-supervised throughout, with no human labels and no pose annotations at any stage.
 
Specializing such a model by simply training it longer on human data does not work, and the two reasons organize this section. The dense features that make the family attractive collapse during adaptation, silently enough that the training objective never reveals it. And the standard masked objective spends the adaptation budget on a copy task, taxing exactly the temporal representation a video JEPA exists for. \hjepa{} therefore keeps the architecture and the initialization, whose abilities are the asset being specialized, and makes two critical changes, both at the masking and anchoring level rather than the architecture level: an anchored context stream that pins dense perception to the initialization, and a forecasting mask family that specializes the temporal representation without taxing it. A third decision is an omission, no person-level masking and no pose input, and a fourth is optional, a two-phase schedule for the start of training.

\subsection{Anchored Continued Pretraining}

Given a video $x$ tokenized into a $(T',H',W')$ grid of spatiotemporal patches, a student encoder $f_\theta$ sees only unmasked (context) tokens and a predictor $g_\phi$ predicts the latent representation of masked tokens, supervised by an EMA teacher $f_{\bar\theta}$.
A context loss additionally supervises predictions at visible positions with multi-layer targets, the mechanism behind V-JEPA~2.1's dense features:
\begin{equation}
\mathcal{L} \;=\; \underbrace{\big\lVert \hat z_{\mathrm{msk}} -
\mathrm{sg}\,[f_{\bar\theta}(x)]_{\mathrm{msk}} \big\rVert_1}_{\text{masked prediction}}
\;+\; \lambda\,
\underbrace{\big\lVert \hat z_{\mathrm{ctx}} -
\mathrm{sg}\,[\,A(x)\,]_{\mathrm{ctx}} \big\rVert_1}_{\text{context (dense) loss}},
\qquad \hat z = g_\phi(f_\theta(x_{\mathrm{ctx}})),
\label{eq:objective}
\end{equation}
where $\mathrm{sg}$ is stop-gradient and $A$ is the context-target network.
Stock training uses $A = f_{\bar\theta}$, which lets the dense targets drift with the student.
We instead set $A = f_0$, a frozen copy of the initialization, so the dense stream is distilled from the one network whose dense quality we want to keep. An image branch trains the same objective on LUPerson-T person crops~\citep{shao2023unified} rendered as one-frame clips, carrying full-body appearance at a scale video cannot.

\subsection{Forecasting Masks, and Mixing the Families}
Standard multiblock masks repeat a spatial block across every frame.
The visible context then always contains the same spatial surround at other instants, so the shortest path to a correct prediction is to copy appearance across space and time.
Under continued pretraining this is destructive rather than merely uninformative, because the adaptation budget is spent strengthening a copy task rather than a model of motion, at a cost we measure on both action recognition and person identity.
Our recipe replaces the blocks with a pure past-to-future split: the context is the first half of the tubelets and the target the second half,
\begin{equation}
\mathcal{L} \;=\; \big\lVert\, g_\phi\big(f_\theta(x_{t < T/2})\big)
- \mathrm{sg}\,[f_{\bar\theta}(x)]_{t \geq T/2} \,\big\rVert_1
\;+\; \lambda\, \big\lVert \hat z_{\mathrm{ctx}} -
\mathrm{sg}\,[f_0(x)]_{\mathrm{ctx}} \big\rVert_1.
\label{eq:forecast}
\end{equation}
The first term now demands a continuation rather than an interpolation, since nothing about the future is visible and the only route to the target is a model of how the scene evolves, while the second term is unchanged from Eq.~\ref{eq:objective} and keeps the dense stream anchored throughout.
The design principle is that the temporal structure of what is hidden matters more than its spatial placement on the person, which is why the recipe uses no human prior in the mask and why we test the natural alternative, part-guided masking~\citep{yuan2023hap}, as a control rather than adopting it.

Because the two families sit at opposite ends of an appearance-versus-dynamics trade-off, we also train per-batch mixtures.
At step $i$ a batch draws the forecasting family with probability $p_i$ and the block family otherwise; $p_i$ is constant or follows a two-phase schedule (forecasting first, blocks late).
Mixing acts across batches, never within one, because batched training requires a fixed masked-token count per batch; the family draw is seeded by the step index so all data workers agree.
The two ends of this sweep are exactly the block recipe and the forecasting recipe, so a single knob traverses the whole space between them.
The released recipe is the pure forecasting endpoint of Eq.~\ref{eq:forecast}; mixing ships as a configuration knob rather than as part of the objective.

\section{Experiments}
\begin{figure}[t]
\centering
\includegraphics[width=\linewidth]{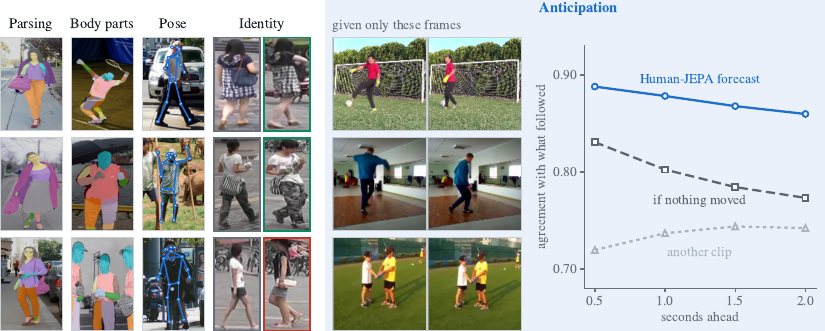}
\caption{\textbf{One frozen checkpoint covers the human perception tasks and anticipates past what it has seen.} All outputs are from this paper's probes on the frozen seed-1 encoder: parsing and surface-part maps, COCO keypoints, and the top Market-1501 match (green frames carry the query identity). For anticipation the released predictor forecasts the latents of the unseen second half of held-out clips; the curve is cosine agreement with the true continuation over 20 clips, above the static reference at every horizon out to two seconds.}
\label{fig:tasks}
\end{figure}

The experiments establish four claims: frozen-probe leadership on pose and identity, a released bundle that anticipates better than the base's, causal attribution of every design decision, and quantified costs. Figure~\ref{fig:tasks} shows what a single frozen checkpoint produces across this range.
 
\subsection{Experimental Setup}
\paragraph{Data.} A person detector inspects four sampled frames of each of 536{,}699 Kinetics-700 clips~\citep{carreira2019short}; a loose gate keeps candidates and records the keypoints of every detected person in a per-shard manifest; the final full-body gate then selects 164{,}431 clips (30.6 percent) by replaying thresholds over the manifests. A second offline gate over the same manifests keeps 82{,}696 clips where at least two solid people are spatially engaged, of which 24{,}524 were recovered from the source archives after the full-body pass had pruned them, again without any re-detection. AIST++~\citep{li2021ai} adds dance motion at 5 percent weight, and the image branch trains on 958k LUPerson-T person crops~\citep{shao2023unified}; video sampling weights are 0.6 full-body, 0.35 interaction, 0.05 AIST++. Every evaluation subset was frozen before any model saw it, the released manifests distribute clip lists and gate parameters rather than video, and the corpus is Kinetics-bounded by construction.
 
\paragraph{Comparison Method.} We compare against the strongest pixel-anchored human specialists, Sapiens2-0.8B~\citep{khirodkar2026sapiens} and HAP ViT-B~\citep{yuan2023hap}, the general contrastive encoder DINOv3-L~\citep{simeoni2025dinov3} as the capacity-matched dense reference, and the frozen V-JEPA~2.1-L~\citep{mur2026v} that we initialize from. The Sapiens2 comparison is capacity-unmatched in its favor (0.8B against our 0.3B). Image baselines process frames independently and have no native video-action probe, so their action and anticipation cells print \na{} rather than a per-frame approximation.
 
\paragraph{Protocol.} Every number comes from a frozen backbone under a probe protocol identical across models: the same head, training budget, frozen evaluation subsets, and feature interface, with no number copied from another paper, so absolute values are not comparable to published fine-tuned results. Dense parsing uses ATR~\citep{liang2015deep} with linear and DPT~\citep{ranftl2021vision} heads at 512 pixels, surface correspondence uses DensePose~\citep{guler2018densepose} with a DPT head, pose uses a COCO keypoint probe~\citep{lin2014microsoft} reporting AP, and action uses attentive probes on NTU RGB-D 120~\citep{liu2019ntu}, with its 26-class two-person mutual slice scored separately, and on the Kinetics-700 validation split, whose probe head trains on an uncurated subset so our own curation cannot bias the column. ReID uses a frozen BNNeck probe on Market-1501: the concatenated token means of four evenly spaced encoder depths on 256 by 128 crops feed a 512-dimensional reduction head trained with label-smoothed cross-entropy and batch-hard triplet under PK sampling, with cosine retrieval, identical for every backbone. One honesty note: both the base model and ours saw Kinetics video during pretraining, so the K700 column measures general-action retention, on which frozen top-1 drops from 54.7 to 51.8, a cost we report rather than a preserved property.
 
\paragraph{Implementation Detail.} Every arm continues pretraining from the released V-JEPA~2.1-L for 200 epochs (60k steps) at global batch 192 video clips plus 576 person crops, in bfloat16 on 4 GPUs, at learning rate $10^{-4}$ with warmup and cosine decay, gradient clipping at 1.0, and a pre-backward loss-spike guard that rejects anomalous batches before they reach the optimizer. Video input is 16 frames at 4 fps and 256 pixels; the forecast mask keeps the first half of the tubelets, and the block family uses standard multiblock geometry. Pilot arms use the identical recipe at 17 or 34 epochs, and the released pair trains single-phase, so the two-phase start enters only as an ablation.
 
\subsection{Results of Perception}
\paragraph{Overall profile.} Table~\ref{tab:main} sizes both halves of the trade. Against the base, averaged over two pretraining seeds, the recipe buys 0.6 AP of pose and 2.7 mAP of ReID and pays 0.9 points of dense parsing, 1.9 of DensePose, a residual 2.7 point action tax on NTU-120 (2.1 on the mutual slice), well under the block-mask recipe's 4.8 but not zero, and 2.9 points of general action on K700. Against the image specialists, the family position does most of the work: the frozen base already leads Sapiens2-0.8B on pose by 2.3 AP at 2.7 times fewer parameters, and the specialization extends that lead to 2.9. Dense parsing is conceded (2.3 points of ATR to Sapiens2), and ReID is the one column specialization flips outright. The two-seed spread is larger than our pilots predicted; we report it on every headline cell, defend the mean rather than the better seed, and quantify the schedule that removes it in the ablation study. The conceded dense gap resists every anchor, auxiliary, and capacity repair we tested, so we treat it as a division of labor rather than a deficit to chase: pixel-anchored single-image pretraining owns static appearance at high resolution, and latent video prediction owns dynamics.
 
\begin{table}[h]\centering\small
\caption{\hjepa{} leads the human specialists on pose and identity while conceding dense parsing and paying on both action columns. Frozen probes measured by us under one protocol; \textbf{bold} best, \underline{underline} second best; $\pm$ is the two-seed spread (ReID is seed 1); \na{} marks image models, which have no video-action probe.}
\label{tab:main}
\begin{tabular}{lccccccc}
\toprule
& & ATR & DensePose & Pose & NTU & Mutual & ReID \\
Model & Params & mIoU $\uparrow$ & mIoU $\uparrow$ & AP $\uparrow$ & top-1 $\uparrow$ & top-1 $\uparrow$ & mAP $\uparrow$ \\
\midrule
\multicolumn{8}{l}{\emph{Human-centric image encoders}} \\
Sapiens2-0.8B & 0.8B & \best{.767} & \best{.666} & .591 & \na & \na & .4522 \\
DINOv3-L & 0.3B & \underline{.766} & \underline{.661} & .564 & \na & \na & \best{.5148} \\
HAP ViT-B & 0.09B & .685 & .492 & .490 & \na & \na & \na \\
\midrule
\multicolumn{8}{l}{\emph{Video JEPA}} \\
V-JEPA 2.1-L (base) & 0.3B & .748 & .649 & \underline{.614} & \best{70.8} & \best{79.6} & .4370 \\
\midrule
\hjepa{} (ours) & 0.3B & .739{\scriptsize$\pm$.004} &
.630{\scriptsize$\pm$.004} & \best{.620{\scriptsize$\pm$.003}} &
\underline{68.2}{\scriptsize$\pm$2.3} & \underline{77.5}{\scriptsize$\pm$1.6} &
\underline{.4635} \\
\bottomrule
\end{tabular}
\end{table}
 
\paragraph{Person re-identification.} ReID is the largest dividend of the specialization and the only column where a deficit against the pixel-anchored specialist turns into a lead. Table~\ref{tab:reid} gives the result: \hjepa{} beats the pixel-anchored specialist (.4635 versus .4522 mAP) and improves 2.7 mAP over its own frozen base, so the specialization helped rather than merely survived, and the identity signal originates in the LUPerson-T image branch, a capability no video self-supervised baseline reports. DINOv3 keeps the overall lead, which we report without qualification: contrastive objectives are instance-discriminative by construction, and the honest claim is best human-video model, not best model. The sharpest row is the last one: the block-mask variant of our own recipe, identical in data, anchor, and budget, collapses to .2666 mAP, seventeen points below the base it started from, while the forecasting recipe gains. The action tax and the ReID collapse are the same phenomenon observed on two task families that share nothing but the encoder.
 
\begin{table}[h]\centering\small
\caption{The mask family decides the identity cost: forecasting gains 2.7 mAP over the frozen base and passes the human specialist, while block masks lose 17 mAP from the same base, data, and budget. Frozen BNNeck probe on Market-1501, identical head and budget per backbone; \textbf{bold} best, \underline{underline} second best.}
\label{tab:reid}
\begin{tabular}{lccc}
\toprule
Model & mAP $\uparrow$ & Rank-1 $\uparrow$ & Rank-5 $\uparrow$ \\
\midrule
DINOv3-L & \best{.5148} & \best{.7586} & \best{.9074} \\
Sapiens2-0.8B & .4522 & \underline{.7055} & .8744 \\
V-JEPA 2.1-L (base) & .4370 & .6974 & .8702 \\
\midrule
\hjepa{} (ours) & \underline{.4635} & .7043 & \underline{.8798} \\
\quad block-mask variant (A1) & .2666 & .5235 & .7271 \\
\bottomrule
\end{tabular}
\end{table}
 
\subsection{Results of Anticipation}
A video JEPA is released as an encoder plus a predictor, and a user who anticipates deploys the bundle, so we evaluate the bundle directly: NTU-120 early action, where the probe observes only the first half of each clip, with the same frozen attentive-probe protocol in every cell and two feature interfaces per model, encoder only and encoder plus shipped head, whose predictions for the unobserved window the probe attends over. Table~\ref{tab:anticip} contains one expected result and one unexpected one. Expected: our encoder anticipates better than the base encoder ($+1.02$), and the full released bundle is $+4.06$ over the base's released bundle. Unexpected, and to our knowledge unreported: \emph{the base's own released head hurts it}, costing 2.98 points against its encoder-only reading, consistent with its upstream training as a cross-space distillation head rather than a same-space forecaster. Our head, trained by forecasting in its own representation space, is the only released head in the comparison that does no harm ($+0.06$). We state the boundary of the claim plainly: at this observation fraction the rollout adds essentially nothing over our encoder alone, so the value demonstrated is representation consistency, not extra anticipatory signal, and a pre-registered variable-horizon test, reported in the ablation study, does not change that verdict at shorter horizons.
 
The released head can also be read directly, without a probe in the way. On 20 held-out Kinetics-700 validation clips the encoder sees the first half of each clip and the predictor is asked for the latents of the second half: its predictions agree with the latents of the frames that actually followed at 0.873 cosine, against 0.798 for holding the last observed latent fixed and 0.735 for scoring the same predictions against a different clip's future. The margin over the static reference holds on every clip and widens with distance, from 0.057 half a second ahead to 0.087 two seconds ahead, and the forecast retrieves the exact moment it was asked to predict 40 percent of the time against a 0.6 percent chance rate. That is the signature of a forecast rather than a copy.
 
\begin{table}[h]\centering\small
\caption{The shipped head is not neutral: the base's released head costs it 2.98 points, while ours is the only one that does no harm. NTU-120 early action with the probe observing the first half of each clip.}
\label{tab:anticip}
\begin{tabular}{lccc}
\toprule
Model & Encoder only & Encoder $+$ shipped head & Effect of the head \\
 & top-1 $\uparrow$ & top-1 $\uparrow$ & $\Delta$ top-1 \\
\midrule
V-JEPA 2.1-L (base) & 78.91 & 75.93 & $-2.98$ \\
\midrule
\hjepa{} (ours) & \underline{79.93} & \best{79.99} & $+0.06$ \\
\bottomrule
\end{tabular}
\end{table}
 
\subsection{Ablation Study}
% Every design decision is tested under matched controls: same corpus, same recipe, one change per arm.
 
\paragraph{The mask family sets the cost, on two unrelated task families.} At a matched 34-epoch pilot horizon, replacing block masks with the pure past-to-future split and touching nothing else moves action by $+8.6$ NTU points overall and $+7.0$ on the mutual slice, at no dense or pose cost. At full scale (Table~\ref{tab:mutual}), every block-masked arm pays about five points of action, spread evenly across two-person and single-person classes, while the forecasting recipe returns to within noise of the base (70.4 versus 70.8; 79.0 versus 79.6 mutual). The same decision controls ReID, a task family the recipe never saw: block masks collapse it by seventeen mAP while forecasting gains (Table~\ref{tab:reid}). The tax is a mask-family artifact, not a data or anchor effect: block inpainting spends the adaptation budget on a temporally local copy task, and forecasting keeps the pressure to model dynamics.
 
\begin{table}[h]\centering\small
\caption{Forecasting returns action to the base level, block masks pay a uniform five-point tax, and whole-person deletion collapses at scale. Frozen NTU-120 probe at full scale; mutual is the 26 two-person classes.}
\label{tab:mutual}
\begin{tabular}{lccc}
\toprule
Model & NTU top-1 $\uparrow$ & mutual $\uparrow$ & non-mutual $\uparrow$ \\
\midrule
A0: V-JEPA 2.1-L (frozen) & \best{70.8} & \best{79.6} & \best{68.1} \\
A1: anchored block-mask adaptation & 65.6 & 74.7 & 62.8 \\
whole-person deletion (200 ep) & 45.5 & 57.0 & 42.0 \\
\midrule
\hjepa{} & \underline{70.4} & \underline{79.0} & \underline{67.7} \\
\bottomrule
\end{tabular}
\end{table}
 
\paragraph{Person-level objectives fail, with a causal probe as judge.} Table~\ref{tab:objective} isolates the omitted person-level family at matched pilot scale. Masking whole people costs dense parsing (0.7357, below the random-mask band of $0.7423 \pm 0.0010$), and the cost is the mask, not the data, since a random-mask control on the same blended interaction corpus stays inside the band (0.7418). Deletion does not teach partner reading (mutual 56.4 versus 56.5), because deleting a person from all of space-time removes them from the target side of every token the loss touches, so two-person co-motion is never a training signal, and the damage compounds with horizon: at 200 epochs dense falls to 0.6996, pose to 0.5403, and action to 45.5 (Table~\ref{tab:mutual}). Any pose input makes it worse: a four-anchor skeleton trajectory scores 52.9 and even a single-anchor identity skeleton 51.5 against the 56.5 unconditioned control, because a masked person's latent target is dominated by body configuration, so the skeleton supplies most of the answer. Person-level forecasting, the Causal-JEPA-style intervention, forfeits the scene-level gain (57.8 versus 69.7) by leaving part of the future visible, reopening the temporal copy leak.
 
The judge throughout is the partner-ablation probe: on 256 held-out two-person clips per checkpoint, it predicts the masked person's tokens with the partner visible and again with the partner's exact token support removed from the context, and reads the relative error increase as a direct causal measure of partner use. Across nine designs, two representation levels (token grid and person slots), and three loss families (point L1, per-person L1, contrastive), partner use never exceeds 0.72 percent on wild clips, and a person-slot InfoNCE arm that solves its retrieval task emphatically (78 percent over roughly 768 candidates, chance 0.13 percent) shows the \emph{least} partner use of any arm, so benchmark-shaped proxies dissociate completely from relational learning. Calibration on scripted NTU mutual validation clips, which no wild-trained arm ever saw, shows the ceiling is a property of the data: given coupling, our scene-forecasting model reads the largest partner use of any wild-trained arm (2.02 percent), while supervised mutual-action co-training suppresses partner use to the floor (0.37 percent) even as it solves its training task. The probe, not any benchmark, is the instrument that makes these distinctions visible.
 
\begin{table}[h]\centering\small
\caption{Making people the unit of prediction never pays: every person-level variant sits at or below the random-mask band on dense parsing and at or below its matched control on interaction, and only scene-level forecasting lifts interaction. Matched-recipe pilots; rows compare within a horizon block; \textbf{bold} best mutual; \na{} marks unprobed cells.}
\label{tab:objective}
\begin{tabular}{lccc}
\toprule
Masking & Dense mIoU $\uparrow$ & Pose AP $\uparrow$ & Mutual top-1 $\uparrow$ \\
\midrule
random multiblock (default) & $0.7423 \pm 0.0010$ & \na & 56.5 \\
random multiblock, blended corpus & 0.7418 & 0.6185 & 55.8 \\
part tubes (HAP-style control) & 0.7357 & \na & \na \\
whole person, $|M|{=}1$ & 0.7357 & 0.6132 & 56.4 \\
whole person $+$ trajectory prior & 0.7363 & 0.6105 & 52.9 \\
whole person $+$ traj.\ prior, $|M|{=}2$ & 0.7354 & 0.6105 & 54.4 \\
whole person $+$ identity prior & 0.7340 & 0.6132 & 51.5 \\
\midrule
random multiblock, blended (34 ep) & 0.7439 & 0.6204 & 62.7 \\
whole person, $|M|{=}1$ (34 ep) & 0.7391 & 0.6224 & 58.5 \\
whole person, $|M|{=}1$ (200 ep) & 0.6996 & 0.5403 & 57.0 \\
\midrule
scene forecasting, no person mask (34 ep) & 0.7413 & 0.6223 & \best{69.7} \\
person forecasting, $t_0$ anchor (34 ep) & 0.7394 & 0.6223 & 57.8 \\
\bottomrule
\end{tabular}
\end{table}
 
\begin{figure}[t]
\centering
\includegraphics[width=0.8\linewidth]{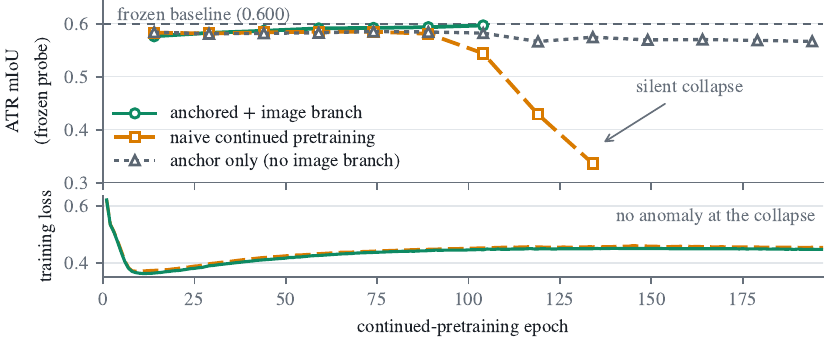}
\vspace{-2mm}
\caption{\textbf{The collapse is silent, and the two preservers play different roles.} Top: frozen ATR linear canaries at matched epochs; the full recipe (green) rises to the frozen baseline, the naive control (orange) collapses after 90 epochs, and the anchor-only run (gray) never collapses but plateaus. Bottom: the training losses, indistinguishable and anomaly-free.}
\label{fig:collapse}
\vspace{-4mm}
\end{figure}

\paragraph{The collapse is silent and selective, and the anchors are complementary.} Naive continued pretraining ($A = f_{\bar\theta}$, video only; row C of Table~\ref{tab:arms}) destroys dense human perception, pose AP falling from 0.614 to 0.110 and DensePose from 0.649 to 0.353, while the training loss decreases throughout and clip-level action drops only four points. 
Matched-epoch probes (Figure~\ref{fig:collapse}) show the control indistinguishable from the anchored run for 90 epochs before collapsing over the next 40, with nothing anomalous in any trainer-visible signal, and the collapse is selective for appearance: under a DPT readout the control's body and geometry classes recover while garment and appearance classes remain collapsed. Counterintuitively, \emph{lower} learning rates make it worse (canary 0.520 at lr $10^{-4}$ versus 0.328 at $2{\times}10^{-5}$, both unanchored), and within each family the lower-rate run reaches a lower final training loss and a worse canary: the optimizer succeeds at the wrong thing, drifting toward features that are easy to predict and less informative. Our reading is that the freshly initialized predictor converges slowly at low learning rate and feeds structured but wrong gradients into the encoder for longer; anchoring works at both rates because it removes the easy basin rather than racing the descent into it, and the same mechanism motivates the two-phase schedule. Table~\ref{tab:arms} then completes a two-by-two factorization of the two candidate preservers. With neither, the model collapses (C). The image branch alone holds the frozen baseline and matches the anchored arm's pose gain (C$_{\mathrm{img}}$), despite training on \emph{uncurated} video, which also shows curation is not the active ingredient at matched objective, with the caveat that uncurated Kinetics is still human-dominated. The anchor alone never collapses, its canary holding a plateau through the exact window where C loses 25 points, but it erodes slowly to 0.709 dense and 0.554 pose, concentrating on garment and appearance classes, the same selectivity in slow motion. The verdict is asymmetric and the mechanisms complementary: the anchor removes the drifting context target that causes the collapse, and the image stream continually re-grounds the appearance features that video-only prediction lets fade, so \hjepa{} uses both.
 
\begin{table}[h]\centering\small
\caption{Naive continued pretraining collapses dense perception while its training loss improves, and each ingredient recovers a specific part of it. Full-scale arms under identical frozen probes; \collapsed{gray} marks collapsed cells, \na{} unprobed metrics.}
\label{tab:arms}
\resizebox{\linewidth}{!}{
\begin{tabular}{lcccccc}
\toprule
Arm & Params & ATR-lin $\uparrow$ & ATR-DPT $\uparrow$ & DensePose $\uparrow$ & Pose AP $\uparrow$ & NTU $\uparrow$ \\
\midrule
A0: V-JEPA 2.1-L (frozen) & 0.3B & .6230 & .7479 & .6490 & .6142 & 70.8 \\
C: naive continued PT (uncurated) & 0.3B & \collapsed{.2947} &
\collapsed{.5907} & \collapsed{.3534} & \collapsed{.1105} & 66.5 \\
C$_{\mathrm{img}}$: C $+$ image branch & 0.3B & .6223 & .7415 & .6367 &
.6262 & 65.4 \\
anchor only (no image branch) & 0.3B & \na & .709 & \na & .554 & 61.9 \\
A1: anchored block-mask adaptation & 0.3B & .6235 & .7452 & .6384 &
\best{.6279} & 65.6 \\
\hjepa{} & 0.3B & .6210 & .7433 & .6333 &
.6233 & 70.4 \\
\bottomrule
\end{tabular}
}
\end{table}
 
\paragraph{You keep what you anchor.} Every continued-pretraining arm pays an NTU action tax under block masks, about five points for the arms with an image branch or naive objective and more still for the anchor-only arm (61.9), because the context anchor pins only the dense spatial features it supervises while temporal dynamics follow the moving teacher and drift with the data distribution. This is precisely the opening for the masking change: rather than anchoring a temporal stream, which would freeze what we want to improve, the forecasting split redirects the \emph{masked} stream to demand temporal prediction, and Table~\ref{tab:mutual} shows this recovers the tax nearly in full.
 
\paragraph{No repair closes the dense gap.} Treating the context target $A$ as a design variable at matched pilot scale: a re-layered self-anchor does not help (0.739); a cross-model DINOv3 anchor routed onto the predictor leaves dense unchanged while dragging pose toward DINOv3's weakness (0.737/0.595); routed onto the encoder it collapses the representation at every weight tried, since latent-predictive and contrastive dense spaces are incompatible; and an auxiliary pixel decoder preserves but does not compound with a longer schedule. Capacity does not close it either: under identical frozen probes the 2B-parameter V-JEPA~2.1-G, seven times our model, gains only 0.006 ATR-DPT and 0.011 DensePose mIoU and remains below Sapiens2-0.8B, while its pose advantage widens. Across self, cross-model, pixel, and capacity repairs the dense metric plateaus near 0.74 against the pixel-anchored references near 0.767, which is why we treat the gap as a division of labor rather than a deficit to chase.
 
\paragraph{Knobs and nulls.} Mixing the two mask families per batch traverses a clean monotone trade-off at 17 epochs: pose and action rise with the forecast probability while dense parsing and DensePose ease down, and no interior point dominates both endpoints, so the pure forecasting objective of Eq.~\ref{eq:forecast} stands and mixing ships as a configuration knob that maps the appearance-versus-dynamics frontier. The two-phase schedule lifts the collapse-sensitive linear canary by 1.1 points at pilot scale, exactly where the fresh-predictor mechanism predicts; at 200 epochs the canary gain does not compound into higher means, and what the schedule buys instead is reproducibility, shrinking the released pair's dense and pose spread roughly tenfold and halving the action spread at a mean cost of 0.5 AP of pose and 1.1 mAP of ReID, so we release the higher-mean single-phase pair and document the schedule as the variance control. Training the predictor across observed fractions $\{1/4, 3/8, 1/2\}$ posts the best dense and pose of any 17-epoch arm yet fails its pre-registered horizon test, leading its fixed-horizon control by only 0.26 points at a held-out quarter-clip fraction against a bar of $+2$, so it ships as a zero-cost knob and no horizon-generalization claim is made. A motion-guided variant that reweights the forecast loss per token by frame-difference magnitude is a null despite provably shifting the objective, its training loss holding 3.3 percent above the control throughout an identical data order: no probe moves by more than 0.1, because frame-difference weights are barely human-selective on curated clips. The pattern is consistent: changing what the encoder \emph{sees} moves the capability profile, while reweighting errors over the same targets does not. One further negative worth stating: a 17-epoch schedule control already matches the 200-epoch model on ReID (.4634 versus .4635 mAP), so identity features saturate early under forecasting masks and long budgets pay elsewhere.

\section{Conclusion}
\hjepa{} is, to our knowledge, the first human-centric vision model that perceives the present and anticipates the immediate future. Two findings made it possible and should transfer to anyone specializing a video JEPA: continued pretraining silently collapses dense perception through target drift, prevented by anchoring dense targets to the frozen initialization with an image branch as the complementary preserver, and the specialization tax lives in the mask family, removed by forecasting rather than inpainting. The natural next step, making people the unit of prediction, is falsified in all nine forms we constructed by a causal partner-ablation probe that ships with the model. Pixel-anchored pretraining owns static appearance at high resolution; \hjepa{} owns people in time.

\bibliography{reference}
\bibliographystyle{iclr2027_conference}

% \appendix
% \section{Appendix}

\end{document}